# Pilot Comparative Study of Different Deep Features for Palmprint Identification in Low-Quality Images

A.S. Tarawneh[1], D. Chetverikov[1,2] and A.B. Hassanat[3]

[1] Eötvös Loránd University, Budapest, Hungary
[2] Institute for Computer Science and Control, Budapest, Hungary
[3] Mutah University, Karak, Jordan

**Abstract**

*Deep Convolutional Neural Networks (CNNs) are widespread, efficient tools of visual recognition. In this paper, we present a comparative study of three popular pre-trained CNN models: AlexNet, VGG-16 and VGG-19. We address the problem of palmprint identification in low-quality imagery and apply Support Vector Machines (SVMs) with all of the compared models. For the comparison, we use the MOHI palmprint image database whose images are characterized by low contrast, shadows, and varying illumination, scale, translation and rotation. Another, high-quality database called COEP is also considered to study the recognition gap between high-quality and low-quality imagery. Our experiments show that the deeper pre-trained CNN models, e.g., VGG-16 and VGG-19, tend to extract highly distinguishable features that recognize low-quality palmprints more efficiently than the less deep networks such as AlexNet. Furthermore, our experiments on the two databases using various models demonstrate that the features extracted from lower-level fully connected layers provide higher recognition rates than higher-layer features. Our results indicate that different pre-trained models can be efficiently used in touchless identification systems with low-quality palmprint images.*

## 1. Introduction

Deep learning has been successfully applied to many computer vision problems including segmentation [3], detection [26] and recognition [20]. Convolutional neural networks (CNNs) and deep learning have been efficiently used to improve the performance of biometric systems [17].

Palmprint identification is an important field due to its unique properties and its usage in crime scene investigation [6]. Palmprint images contain many distinct, specific features which can be used for identification purpose [14]. Researchers investigate robust features that make the identification and authentication more accurate [11].

Recently, there has been growing interest in using CNNs to obtain useful deep features for many tasks of recognition and classification [7, 21]. Most of the available databases for biometrics are relatively small [8], hence using these databases to train a CNN from scratch would increase the risk of overfitting. In addition, training CNN from scratch on a large database takes a long time [23]. To overcome these problems, researchers started to use features extracted from a pre-trained CNN, e.g., AlexNet [12], VGG-Net [22], etc. These networks are trained on a large-scale database such as ImageNet [5].

In this work, three popular pre-trained CNN models, AlexNet, VGG-16 and VGG-19, are compared in their capabilities to solve the problem of palmprint identification in low-quality imagery. The pre-trained models are used to extract the features from the palmprint images. A challenging hand image database is used to study the ability of different pre-trained models to provide deep features for recognition of low-quality palmprints. Another, high-quality hand image database is also processed to compare the recognition accuracies at low and high image qualities. The features are classified using the multi-class Support Vector Machine (SVM) with stochastic gradient descent. The performance is evaluated for features extracted from different layers, databases and at different training rates.

The rest of this paper is organized as follow. Section 2 provides a discussion of related work. The benchmark data used in our experiments is presented in section 3, the pro-

*Tarawneh et al. / Palmprint identification*

posed feature extraction approach in section 4. In section 5, we discuss our experiments and their results. Finally, conclusions and future work are presented in section 6.

## 2. Previous Work

Pre-trained CNNs have been used to solve various computer vision problems. Kumar et al. [13] presented a comparative study of two pre-trained models, AlexNet and VGG-16, in histopathological image classification. They used four similarity measures to compare the extracted feature vectors. However, the comparison did not cover the differences in performance between VGG-16 and VGG-19, as well as features extracted from different fully connected layers.

Minaee et al. [17], published an experimental study of pre-trained VGGNet for iris recognition. The results indicate high performance as the best accuracy was around 94% on a popular iris database.

Numerous methods and approaches have been developed and used for palmprint identification systems. The performance of the palmprint recognition system [4] was evaluated on high-quality palmprint images acquired by a scanner. Wavelet transform and different projection methods were used to analyze the images. Good recognition performance was achieved with low false acceptance (FA) and false rejection (FR) rates.

Kong et al. [11] presented a survey of several palmprint identification systems. The survey covers tools and operations for palmprint recognition including devices used to capture palmprint images, and preprocessing and verification techniques.

Minaee and Abdolrashidi [16] proposed an approach to palmprint identification that uses a combination of discrete cosine and wavelet transforms and employs Principal Component Analysis (PCA) for dimensionality reduction. Their experiments show very high success rates reaching 99.79%-100% on the high-quality palmprint image database PolyU [24, 27]. Wavelet features were also used in the study [15] where fusion of wavelet features and statistical features was employed to create the final feature vector. The same database was used to evaluate the performance of the system, with the success rates up to 99.65%-100%.

The palmprint recognition system introduced in [18] applied a deep convolutional neural network called scattering network. Since the number of scattering features extracted from the image was high, PCA was used to reduce the dimensionality of the feature vectors. Then multi-class support vector machines classifier was applied for classification. The proposed method was also evaluated on the PolyU database, and the accuracy reached 99.95%-100%.

In the above studies, excellent identification results were reached on the PolyU database. This high performance can partially be attributed to the high quality of the database imagery obtained by a palmprint image acquisition device that captures multispectral data under blue, green, red and near-infrared illuminations [24]. In our study, we concentrate on low-quality data, which is more typical for real-world applications.

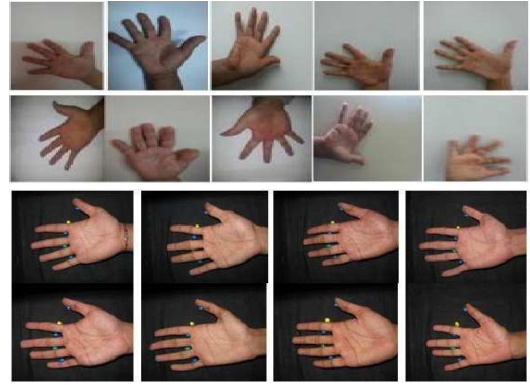

**Figure 1:** *Samples from MOHI (top) and COEP (bottom).*

## 3. Benchmark Data

In contrast to the high-quality PolyU imagery, many non-trivial problems arise when one wants to deal with palmprint identification in practice. Shadows, non-uniform illumination, scale variation, rotation, translation and low contrast should be coped with to develop a robust palmprint recognition system capable of operating in real-world conditions.

Most of the databases used in the literature do not have such problems since the images were captured by high-quality cameras or scanners under predefined conditions, such as strictly fixed position of hands. For these reasons, we decided to work with MOHI [8], a more challenging database where the users could move their hands freely without predefined conditions. The images were captured by a smartphone camera under varying conditions. The database was created for 200 subjects; for each subject, the images were captured in 3 sessions with 5 images in each session. In total, the database contains 3000 images for 200 subjects with 15 images each. It is publicly available at `https://www.mutah.edu.jo/biometrix`.

To compare the performances of pre-trained models on MOHI and a higher-quality imagery, we also used the palmprint database called COEP [1] acquired under much better conditions than MOHI. In particular, the COEP palmprints are shadow-free, distinct shapes against uniform background. They show fine skin details and have fixed orientation and constrained finger positions. Neither of these properties is typical for MOHI. Samples from both databases are demonstrated in figure 1.

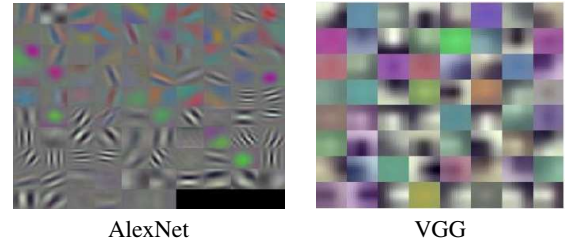

Figure 3: *First-layer weights of AlexNet and VGG.*

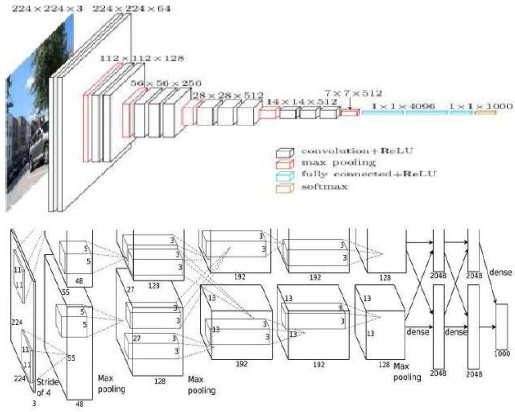

**Figure 2:** *Network architectures of VGG (top) and AlexNet (bottom). Illustrations courtesy of [25] and [9].*

## 4. Feature Extraction

Deep features are image features extracted using pre-trained CNN models such as AlexNet [12] or VGG16 and VGG-19 [22]. We use VGGNet and AlexNet because they have been widely and successfully applied in many applications [2]. Also, we wish to clearly demonstrate the role of network deepness in feature extraction and object recognition. Figure 2 illustrates the architectures of the VGG and AlexNet models, while figure 3 shows the first-layer weights of the models. Both models were trained on the large-scale database ImageNet [5].

The main steps of our feature extraction method are as follows:

1. Segment the input color image into hand and background by the K-means algorithm [10, 19]. To speed up the procedure, reduce image size by 30% prior to segmentation.
2. Enhance the contrast of the hand segment by processing each channel (R,G,B) separately.
3. Calculate the centroid and the major axis of the hand mask. Normalize hand orientation by rotating the hand to make its major axis horizontal.
4. Apply morphological operations to clean the hand mask and remove holes, if any.
5. Obtain the palmprint region of interest (ROI) as the maximal area square within the orientation-normalized hand mask. Use the corresponding square region of the enhanced color hand image to extract features.
6. Resize the ROI image to fit the input layer of each model: $224 \times 224$ pixels for VGG-16 and VGG-19, and $227 \times 227$ pixels for AlexNet.
7. Extract features using each of the three models.

The contrast enhancement as well as the orientation and size normalization steps make the method less sensitive to illumination, orientation and size variations, respectively. The models provide 4096-dimensional feature vectors. Multi-class SVM with stochastic gradient descent is applied to classify the feature vectors. The stochastic gradient descent is used to speed up the high-dimensional classification procedure.

## 5. Experiments and Results

Recall that for performance evaluation we use the databases MOHI (low-quality) and COEP (high-quality). The data from each database is divided into a training set and a test set. The training set is selected randomly from each class. For comprehensive evaluation of the performance, we vary the training set size from 10% to 90% of the data. More specifically, we first use 10% for training and the remaining 90% for testing, then 20% for training and the rest for testing, and so on.

To avoid the bias which may occur because of the randomness in selection of the training set, the experiment was repeated 10 times for each division of the database, then the average of the accuracies was calculated as the final result. This was done for each experiment of the study. We extracted the features from the fully connected (FC) layers number six and seven (FC6 and FC7) of the three models. Figure 4a shows the three plots of the classification accuracy against the training ratio on the MOHI database using the features extracted from FC6. Note that in this case the plots of VGG-16 and VGG-19 are too close for visual separation.

With FC6 features, VGG-16 and VGG-19 provide almost identical performance which is consistently better than that of AlexNet. The accuracy starts increasing from 69% at the training ratio of 10% and gradually grows to reach almost 95.5% at the ratio of 90%. For AlexNet, the accuracy starts from 60% and grows to 93%.

The features extracted from FC7 yield lower accuracies than for those of FC6. For VGG-16 and VGG-19, the accuracies start from about 65% and 62%, respectively, with the maximum around 91.5% using 90% of each class as training data. The AlexNet accuracy starts from 53% with the maximum of 89%. Figure 4b illustrates the performance of the models on MOHI with the features extracted from FC7 for varying training ratios.

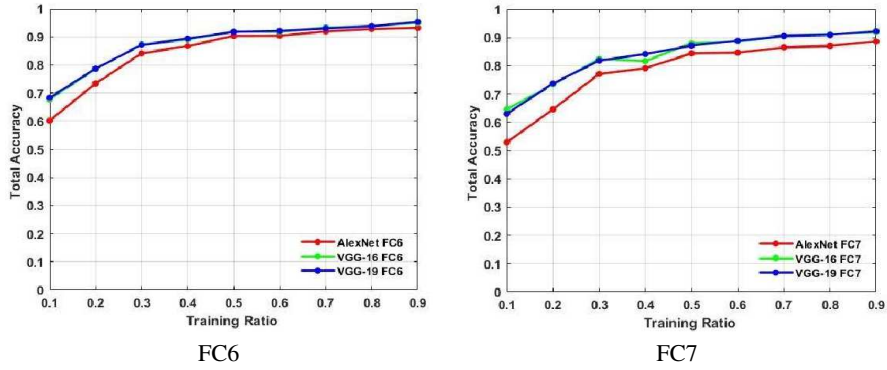

| FC6 | FC7 |

**Figure 4:** *Comparative results on **MOHI** for FC6 and FC7.*

On the MOHI database, FC6 features consistently outperform FC7 ones for all models used in the experiments. The difference in accuracy between FC6 and FC7 is considerable in all cases, especially for AlexNet. The difference slightly decreases with the training ratio, but remains visible even for the largest ratio.

We also conducted the same experiments on the COEP database to evaluate the performance on a higher-quality image database. Figures 5 shows the accuracy plots of the models using features extracted from FC6 and FC7. Here, the accuracies achieved by the models increase fast as the training ratio starts growing, then level off at large values. This is due to the detectable and distinctive features obtained for the higher quality images of COEP.

Nevertheless, the accuracies of FC6 still exceed those of FC7 on COEP, as well. For FC6, the success rates start at almost 67% for AlexNet and VGG-19, and 71% for VGG-16. The success rates grow to approximately 98.5% reaching maxima at training ratios between 70% and 90%. For FC7, features the success rates start at 65% for VGG-16 and AlexNet, and 58% for VGG-19. The largest success rates are almost 98% achieved at the training ratio of 90%. Despite the fact that VGG-19 is deeper than AlexNet and VGG-16, AlexNet and VGG-16 outperform VGG-19 with FC7 features for the training rates between 10% and 50%.

Similarly to the MOHI tests, FC6 features consistently outperform FC7 ones for all models used in the COEP experiments. The difference in accuracy is considerable for VGG-16 and VGG-19, while AlexNet is much less sensitive to the selection of the layer. In all the three cases, the difference between FC6 and FC7 decreases with the training ratio and becomes negligible when the ratio approaches 0.5.

## 6. Conclusion and Outlook

In this study, we conducted several sets of experiments on two palmprint image databases using three pre-trained convolutional neural networks. SVM was used to classify the deep features extracted from the CNN models. Varying training ratios were tested to analyze the performance of the proposed palmprint recognition system and its relation to the number of training samples.

The first set of the experiments aimed at testing the features extracted from the low-quality palmprint images of the MOHI database. Overall, the results show that the features extracted from a deep pre-trained CNN provide better recognition rates than the features of a less deep CNN. VGG-16 and VGG-19 proved to be very efficient on the MOHI database, with the starting accuracy of 69% when 2 images out of 15 were used for training and 13 for testing. The highest accuracy of approximately 95.5% was achieved by both models using 14 images for training. For AlexNet, the corresponding accuracies were 60% and 93%. On MOHI, the lower recognition rates at lower training ratios are partially because of the numerous shadows that make the segmentation less accurate. This resulted in parts of fingers being included in ROI and adding features that did not belong to the palmprint. Another source of errors can be occasional distortions in MOHI images because of skew and slant in hand orientation: sometimes, palm planes are visibly not parallel to the image plane.

Our second set of tests demonstrates the essential difference in performance on high-quality and low-quality palmprint images. On the high-quality palmprint database COEP, all of the tested models perform very similarly at the training ratios 50%-90%. We believe that this is because the COEP features are numerous, distinct and easy to extract, so all of the models can efficiently obtain the informative features. On the low-quality MOHI database, the deeper networks are more accurate in the feature extraction process.

Finally, the third part of our study shows how features extracted from different fully connected layers affect the performance of the system. For all tested models, we used features from two different layers, FC6 and FC7. The results for both databases and various models demonstrate that SVM works better with FC6 features than with FC7 ones.

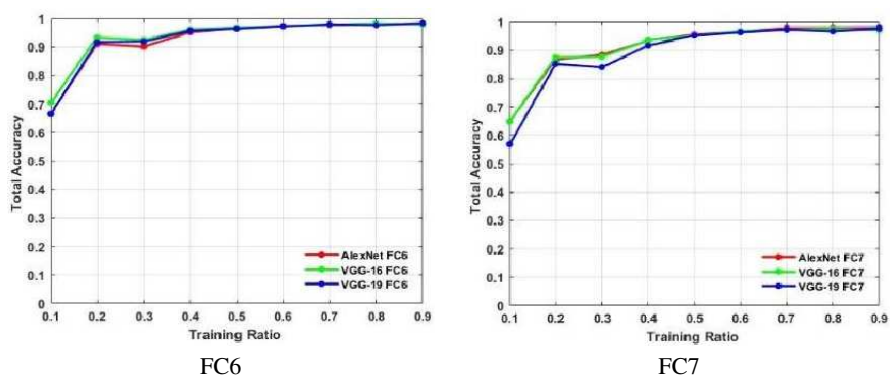

FC6            FC7

**Figure 5:** *Comparative results on **COEP** for FC6 and FC7.*

As future work, we plan to reduce the feature vectors by different dimensionality reduction techniques, such as the Principal Component Analysis (PCA) and the Discrete Wavelet Transform (DWT). Furthermore, we will address the shadow removal and distortion compensation problems in order to enhance hand segmentation and ROI extraction. We also plan to artificially increase the number of images in MOHI by applying geometrical and intensity transformations. Overfitting avoidance, e.g., dropout, and generalization techniques will be used. Finally, we plan to compare low-quality palmprint identification accuracies achieved by pre-trained CNNs and a CNN built and trained from scratch.

**Acknowledgments**

The project is supported in part by the European Union, co-financed by the European Social Fund (EFOP-3.6.3-VEKOP-16-2017-00001).